\documentclass[10pt,twocolumn,letterpaper]{article}

\usepackage[pagenumbers]{cvpr} 
\usepackage{graphicx}
\usepackage{float}
\definecolor{cvprblue}{rgb}{0.21,0.49,0.74}
\usepackage[pagebackref,breaklinks,colorlinks,allcolors=cvprblue]{hyperref}

\def\paperID{11} 
\def\confName{CVPR}
\def\confYear{2025}

\title{Text-Driven Diffusion Model for Sign Language Production}

\author{
Jiayi He~~~~~
Xu Wang~~~~~
Ruobei Zhang~~~~~
Shengeng Tang\thanks{Corresponding author.}~~~~~
Yaxiong Wang~~~~~
Lechao Cheng\\
\textsuperscript{}School of Computer Science and Information Engineering, Hefei University of Technology \\
{\tt\small \{hejy4396,wangxu2002,2024170851\}@mail.hfut.edu.cn, \{tangsg,wangyx,chenglc\}@hfut.edu.cn}
}

\begin{document}
\maketitle
\begin{abstract}
We introduce the hfut-lmc team's solution to the SLRTP Sign Production Challenge. The challenge aims to generate semantically aligned sign language pose sequences from text inputs. To this end, we propose a Text-driven Diffusion Model (TDM) framework. During the training phase, TDM utilizes an encoder to encode text sequences and incorporates them into the diffusion model as conditional input to generate sign pose sequences. To guarantee the high quality and accuracy of the generated pose sequences, we utilize two key loss functions. The joint loss function $\mathcal{L}_{joint}$ is used to precisely measure and minimize the differences between the joint positions of the generated pose sequences and those of the ground truth. Similarly, the bone orientation loss function $\mathcal{L}_{bone}$ is instrumental in ensuring that the orientation of the bones in the generated poses aligns with the actual, correct orientations. In the inference stage, the TDM framework takes on a different yet equally important task. It starts with noisy sequences and, under the strict constraints of the text conditions, gradually refines and generates semantically consistent sign language pose sequences. Our carefully designed framework performs well on the sign language production task, and our solution achieves a BLEU-1 score of $20.17$, placing second in the challenge.

\end{abstract} 
\section{Introduction}
\label{sec:intro}

\begin{figure}[t]
\centering
\includegraphics[width=1.0\linewidth]{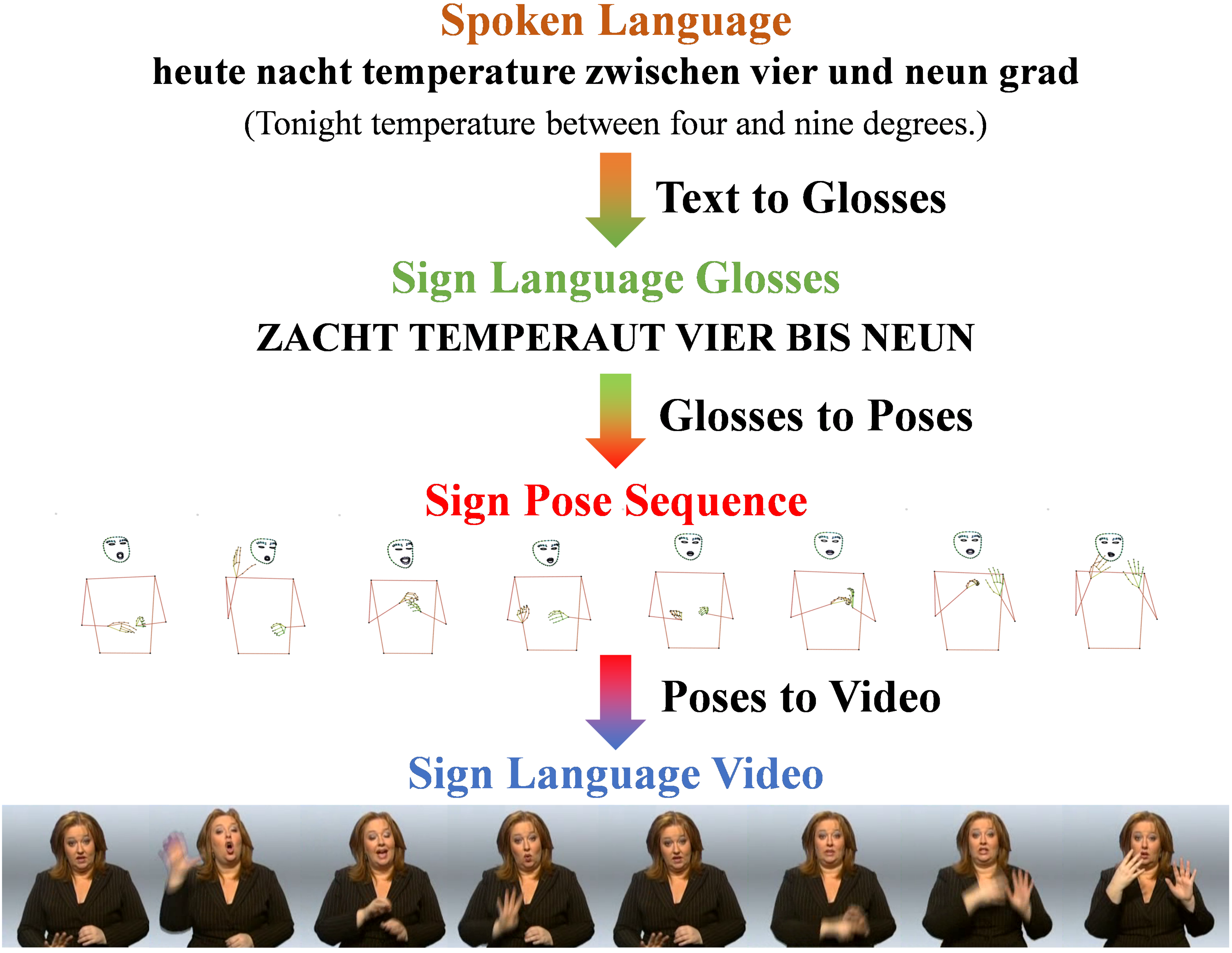}
\caption{The task of generating sign language pose sequences directly from text extremely daunting challenges. Classic pipelines for Sign Language Production: text-to-gloss, gloss-to-pose and pose-to-real video. In this work, we go beyond the commonly used pipeline and focus on the T2P task of generating sign pose sequences directly from text.}
\label{fig: task}
\end{figure}

\begin{figure*}[th]
\centering
\includegraphics[width=1.0\textwidth]{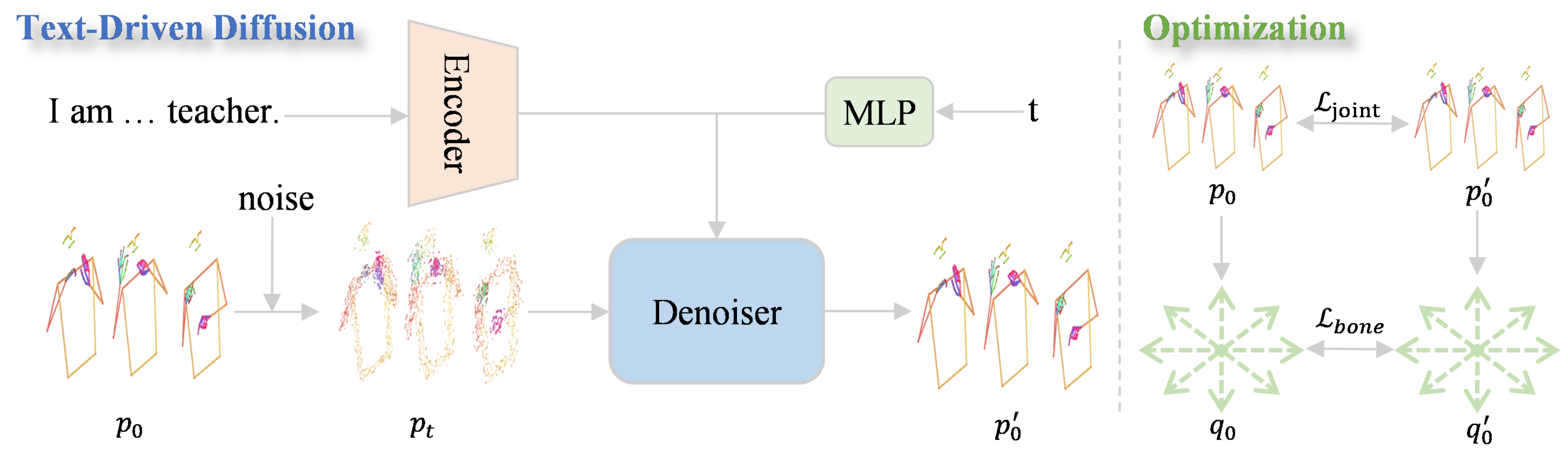}
\caption{Overall framework of our method - TDM. It consists of a text encoder, a denoiser, and a MLP. The text encoder is primarily designed to capture the global semantics of the text sequence. This global semantics is integrated with the time step that has been processed by a MLP to form the relevant condition $g$. We add noise in $t$ steps to the target pose sequence $p_0$ to obtain $p_t$. Then, we send $p_t$ along with the relevant condition $g$ to the denoiser $\mathcal{D}$ to recover the target pose sequence $p_{0}'$ that is free from noise contamination. For the optimization, we employed two main loss functions. The $\mathcal{L}_{joint}$ loss function is primarily used to constrain the joint coordinates, while the $\mathcal{L}_{bone}$ loss function is mainly applied to constrain the direction of bone movement.}
\label{fig: main}
\end{figure*}

Sign language, an essential tool for seamless interaction between key research fields and the deaf community, is at the forefront of a technological revolution that significantly impacts the construction of information accessibility. In the digital age, with information flowing freely in most of society, ensuring the deaf can equally access this knowledge wealth is not only a matter of social justice but also a frontier in academic and technological exploration.

Currently, academic research in sign language technology mainly focuses on three major interrelated directions: Sign Language Translation (SLT), Sign Language Recognition (SLR), and Sign Language Production (SLP). SLT aims to bridge the communication gap between sign language and spoken languages, facilitating effective cross-communication between the deaf and the hearing-impaired. SLR, in contrast, focuses on the automatic recognition of sign language gestures. Although significant progress has been made in SLT~\cite{ye2023cross, yao2023sign, gong2024llms, tang2021graph, guo2019connectionist} and SLR~\cite{hu2023continuous, zuo2023natural, xue2023alleviating}, SLP still poses a complex and urgent challenge. The complexity of SLP lies in the unique grammatical rules of sign language. Sign language is not merely a visual equivalent of spoken language; it is a separate language system with its own grammar, syntax, and semantics. Spoken language depends on auditory sounds and the temporal order of words, while sign language uses spatial positioning and gesture directionality to clearly show the relationships between sentence components. These spatial and gestural features are crucial to sign language grammar. For instance, the hand's position in space can denote the subject, object, or location in a sign language sentence, and the gesture's movement direction can express actions like moving towards or away from the signer.

To address this challenge, existing methods generally adopt a two-step process. First, they convert text into gloss\footnote{Glosses differ from texts, with each gloss corresponding to a sign language.}, a representation form that captures sign specific meaning. Then, sign language pose sequences are generated from this gloss. However, this approach has several shortcomings. The conversion from text to gloss and then to sign language poses adds multiple levels of complexity. It demands in-depth knowledge of both the source text language and sign language grammar, along with complex algorithms for the conversion. This complexity not only increases the computational load but also severely impacts the efficiency of SLP. In practical applications like real-time communication scenarios, this inefficiency can be a significant obstacle. Therefore, developing methods to generate sign language directly from text is not only highly urgent but also of crucial significance. These methods have the potential to revolutionize sign language production, making it more accessible, efficient, and user-friendly for the deaf community.

To this end, we proposed a Text-Driven Diffusion Model (TDM) for sign language production. TDM employs a diffusion model architecture, encoding text sequences via an encoder and integrating them into the diffusion model as conditioning inputs for pose sequences generation. Meanwhile, we use bone orientation loss $\mathcal{L}_{bone}$ to improve the quality of sign language pose sequences. Our main contributions are:

\begin{itemize}
\item We introduce the diffusion model into the T2P task and use the skeleton constraint loss to generate results that are much more semantically rich than the baseline model.

\item In the SLRTP Sign Production Challenge, our solution achieved a BLEU-1 score of 22.17 and won the second place in the competition. The visualization results show that the results generated by our model are very close to the ground truth.
\end{itemize}

\section{Related Work}
\label{sec:related}

\subsection{Sign Language Production}
Early research in SLP mainly employed avatar-based methods~\cite{brock2020learning, karpouzis2007educational}. However, due to the technological limitations that were prevalent at that time, the avatars showed rather jerky movements. For example, when an avatar was programmed to perform a simple walking motion, it would seem to lurch forward in an unnatural and disjointed way. Furthermore, their expressions were extremely rudimentary. In educational applications, an avatar designed to convey learning-related emotions might only be able to display a very general "happy" or "sad" expression, lacking the subtle nuances essential for effective communication. In the generation task, Variational Autoencoders (VAEs)~\cite{kingma2013auto} and Generative Adversarial Networks (GANs)~\cite{Goodfellow_Pouget-Abadie_Mirza_Xu_Warde-Farley_Ozair_Courville_Bengio_2017} showed remarkable performance. Inspired by their performance, a series of methods based on Generative Adversarial Networks (GANs)~\cite{saunders2021continuous, saunders2020adversarial} and Variational Autoencoders (VAEs)~\cite{hwang2021non, xiao2020skeleton} have emerged. Although these methods try to control the mapping from text to visual representation by encoding text information, they often fail to handle complex semantics effectively. Take a sentence with multiple nested clauses and idiomatic expressions as an example. The encoding-based mapping has difficulty in accurately translating the complete depth of meaning within such a sentence into a corresponding visual representation. The Transformer architecture~\cite{tang2022gloss, wang2024} represents a significant advancement. By utilizing an attention mechanism, it can simultaneously capture rich semantic information in the text from various representation subspaces. This allows it to generate more coherent and natural sign language sequences. Nevertheless, sign language presents unique challenges because of its rich vocabulary, complex grammatical structure, and diverse expressions. To overcome these challenges, the diffusion model has been introduced. G2P-DDM~\cite{xie2024g2p} and Sign-IDD~\cite{tang2024signidd} are diffusion-based solutions. These models generate coordinate representations of hand gestures starting from Gaussian noise. By gradually denoising the noise, they can precisely depict the spatial positions and movement trajectories of sign language gestures, holding great potential for a breakthrough in SLP.

\begin{figure}[t]
\centering
\includegraphics[width=0.8\columnwidth]{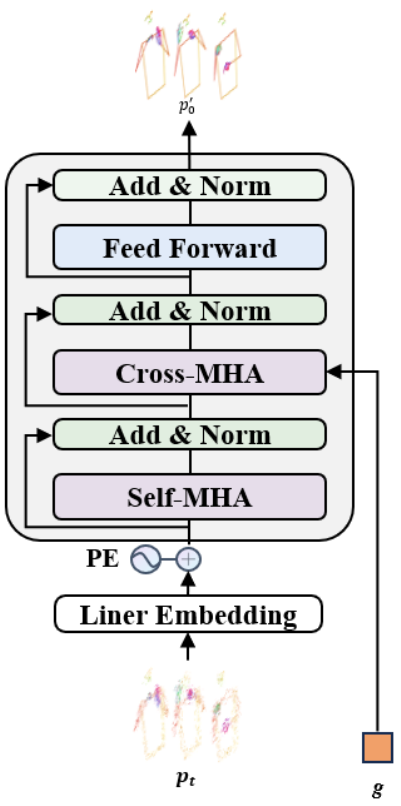}
\caption{Detailed implementation details of our denoiser $\mathcal{D}$.}
\label{fig: Denoiser}
\end{figure}

\subsection{Diffusion Models}
The diffusion model~\cite{Sohl-Dickstein_Weiss_Maheswaranathan_Ganguli_2015} is a sophisticated generative model. It first gradually adds noise to the original data and then, through a reverse denoising process, generates high-quality samples. This unique mechanism has allowed diffusion models to make significant progress in numerous domains. For example, in the field of image generation, diffusion models have achieved remarkable results. For example, DDPM~\cite{Ho_Jain_Abbeel_Berkeley} and DDIM~\cite{Song_Meng_Ermon_2020} have demonstrated that diffusion models are capable of generating highly realistic and diverse images. These models can capture the complex details and semantic information in natural images, thus revolutionizing the field of image generation. Diffusion models have also demonstrated great potential in text generation\cite{li2022diffusion, gong2022diffuseq}. In the area of image inpainting, where diffusion models~\cite{lugmayr2022repaint} have also had an impact. When given an image with missing parts, diffusion-based inpainting algorithms can accurately predict and fill in the missing regions, thereby restoring the image to its original or a plausible state. Moreover, in the field of video generation, diffusion models \cite{ho2022video} and \cite{chen2024videocrafter2} have been able to generate smooth and visually appealing videos while taking into account the continuity of frames. However, despite their extensive success in these areas, the applications of diffusion models in sign language are still relatively limited. Therefore, in this study, we aim to bridge this gap by exploring the potential of propagation models in the SLP.
\section{Methodology}
\label{sec:method}
As shown in Figure~\ref{fig: main}, the TDM mainly comprises three components: 1) the forward process of gradually introducing noise into the 3D pose sequences; 2) the reverse process that recovers the original pose from the noisy pose sequences through denoising; and 3) the denoiser $\mathcal{D}$.

\subsection{Forward Process} \label{Forward Process}
The forward process begins with the target pose sequence $p_{0}$. In this process, noise is added to $p_{0}$ successively in $t$ steps, where $t \in T$. Specifically, at each step, the noise is incorporated in a way that gradually transforms the initial pose sequence into a noisy one. Eventually, this results in the generation of the noisy pose sequence $p_{t}$. The formula is as follows:

\begin{eqnarray}
    p_{t}=\gamma_{t}p_{0}+\sigma_{t}\epsilon,
    \label{eq: Forward Process}
\end{eqnarray}
where $\epsilon$ is sampled from a Gaussian distribution. The parameters $\gamma_{t}$ and $\sigma_{t}$ play crucial roles in determining, respectively, the contribution of the original information in $p_{0}$ and the added noise $\epsilon$. These parameters determine the ratio of the original pose information to the noise in $p_{t}$. Moreover, they satisfy the condition $\gamma_{t}^2+\sigma_{t}^2=1$~\cite{Ho_Jain_Abbeel_Berkeley, Song_Meng_Ermon_2020} , which is a fundamental relationship in this model.


\begin{figure}[t]
\centering
\includegraphics[width=0.8\columnwidth]{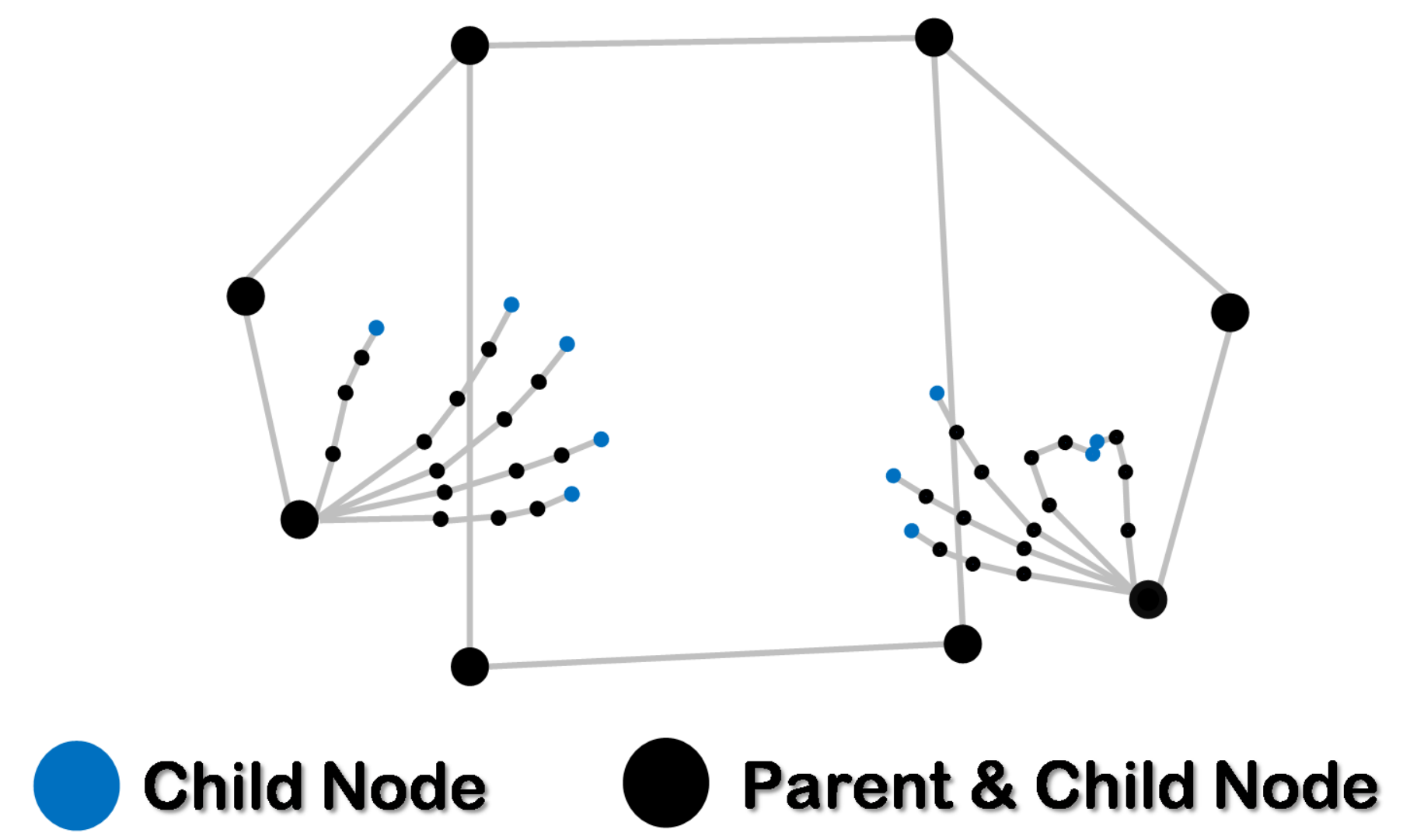}
\caption{Parent-child node partition relationship on the PHOENIX14T dataset.}
\label{fig: Pose}
\end{figure}

\subsection{Reverse Process} \label{Reverse Process}
The training phase in this context is of utmost importance as it endeavors to learn a denoiser $\mathcal{D}$. This denoiser $\mathcal{D}$ differs from the denoiser in common diffusion models~\cite{Song_Meng_Ermon_2020, Ho_Jain_Abbeel_Berkeley} and is mainly used to predict the target pose sequence instead of the noise. The relevant condition $g$ plays a vital role in guiding the denoising process. The denoising operation is expressed by the following formula:
\begin{eqnarray}
p_{0}'=\mathcal{D}(p_t,g),
\label{eq: denoiser}
\end{eqnarray}
where $p_t$ represents the data that is contaminated by noise. 

The denoiser $\mathcal{D}$ takes this noisy data $p_t$ and the relevant condition $g$ as inputs. Through a series of operations and transformations learned during the training process, the denoiser $\mathcal{D}$ aims to generate $p_{0}'$. The value of $p_{0}'$ is the result of the denoising process. It is expected to be a cleaner version of the original data, free from the interference of the added noise. Essentially, the denoiser $\mathcal{D}$ is trained to map the noisy data $p_t$ to the denoised output $p_{0}'$ according to the information provided by the relevant condition $g$. This training process enables the denoiser $\mathcal{D}$ to capture the characteristics of the noise and the relationship between the noisy data and the desired clean data, thus enabling it to effectively perform the noise removal task.

In the inference stage, because the data after the forward process can be well approximated to a Gaussian distribution, we sample noise from the standard Gaussian distribution to initialize the sign language pose sequence $p_T$. By sending the initialized sign language pose sequence $p_T$ and the relevant condition $g$ to the denoiser $\mathcal{D}$, we can predict the sign language pose sequence $\tilde{p}_{0}$ that is free from noise contamination. This inference process can be expressed by Eq.~\ref{eq: denoiser}. Subsequently, $\tilde{p}_{0}$ is utilized to generate the noisy pose sequence $\tilde{p}_{t^{'}}$ at the next time step, which serves as the input of DDIM~\cite{Song_Meng_Ermon_2020}. The formula is as follows:
\begin{eqnarray}
\tilde{p}_{t^{'}} = \gamma_{t^{'}}\tilde{p}_{0}+\sigma_{t^{'}}\epsilon,
\end{eqnarray}
where $t^{'}$ represents the next step after $T$. $\epsilon\sim\mathcal{N}(0, 1)$ is standard Gaussian noise independent of the input $p_T$. Finally, we use $\tilde{p}_{t^{'}}$ as the input to the denoiser $\mathcal{D}$ and re-predict the output $\tilde{p}_{0}$ with the relevant condition $g$. This process will be iterated $i$ times.

\subsection{Denoiser} \label{Denoiser}
To comprehensively and clearly understand the denoiser $\mathcal{D}$, a systematic exploration beginning with its training phase is essential. The implementation details of the denoiser $\mathcal{D}$ are shown in Figure~\ref{fig: Denoiser}. 

It begins with the Linear Embedding Layer (LE), which is a crucial component serving as a bridge between the original noisy sign language pose sequence $p_t$ and the finer dense space. This layer functions to transform the input sequence into a representation that is more suitable for subsequent processing. The formulation of the Linear Embedding Layer is as follows:

\begin{eqnarray}
    p_{u}=W^{p}\cdot p_{t}+b^{p},
    \label{eq: Linear embedding layer}
\end{eqnarray}
where $W^{p}$ represents the weight matrix, which plays a pivotal role in learning the relationships and patterns within the noisy pose sequence. The weight matrix is optimized during the training process to ensure that the transformed representation captures the essential features of the input. The bias term $b^{p}$ is added to introduce a degree of flexibility and prevent the model from being overly restricted. It helps in fine-tuning the output of the linear transformation.

Following the linear embedding operation, the subsequent step is to enhance the temporal characteristics of the transformed sequence. Temporal information holds great significance in sign language analysis because the order and timing of poses carry substantial semantic meaning. To accomplish this goal, we utilize positional encoding (PE). The process of positional encoding is defined by the following equation:
\begin{eqnarray}
    \hat{p_{u}} = p_{u} + PE(n),
    \label{eq: positional encoding}
\end{eqnarray}
where the $PE$ function is a predefined sinusoidal function. This selection of a sinusoidal function is not random; it is specifically designed to offer a smooth and continuous encoding of the relative position $n$ of each frame within the sign language pose sequence. The sinusoidal characteristic of the encoding enables the model to effectively capture both local and global positional information. By incorporating the positional encoding into the linearly embedded sequence $p_{u}$, we are essentially augmenting the representation with information regarding the position of each pose within the sequence. This allows the subsequent layers of the denoiser $\mathcal{D}$ to have a better understanding of the temporal relationships among different poses.

With the linearly embedded and position-encoded sequence $\hat{p_{u}}$ and the relevant condition $g$ in hand, we move on to the final processing stage. This stage utilizes two powerful attention mechanisms: multi-head attention and cross-attention mechanism. The multi-head attention mechanism is a fundamental component of modern neural network architectures, particularly in tasks related to sequential data. It enables the model to simultaneously focus on different parts of the input sequence, thus capturing a more comprehensive range of features. The core of the multi-head attention mechanism is based on the following set of equations:

\begin{eqnarray}
   Attention(Q, K, V)=softmax(\frac{QK^T}{\sqrt{d_{k}}})V,\\
   head_{i} = Attention(QW^{Q}_i, KW^{K}_i, VW^{V}_i),\\
    MHA(Q,K,V) = (head_{1}, \dots, head_{h}) \cdot W.
    \label{eq: multi-head-attention}
\end{eqnarray}

\begin{figure*}[ht]
  \centering
  \includegraphics[width=1\textwidth]{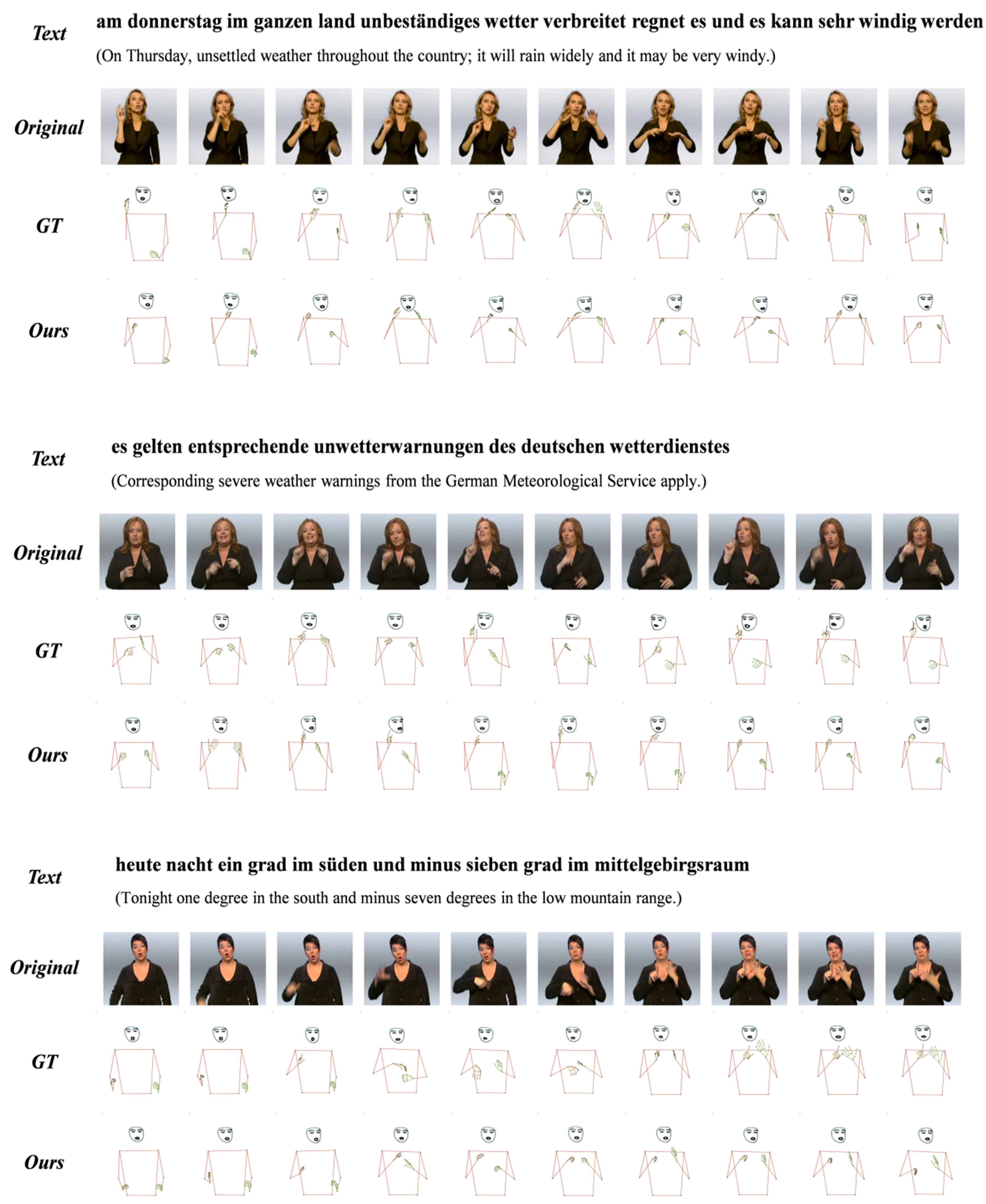}
  \caption{Visualization examples of generated pose sequences on PHOENIX14T. We compare TDM with ground-truth.}
  \label{fig:Visual 1}
\end{figure*}

\begin{figure*}[]
  \centering
  \includegraphics[width=1\textwidth]{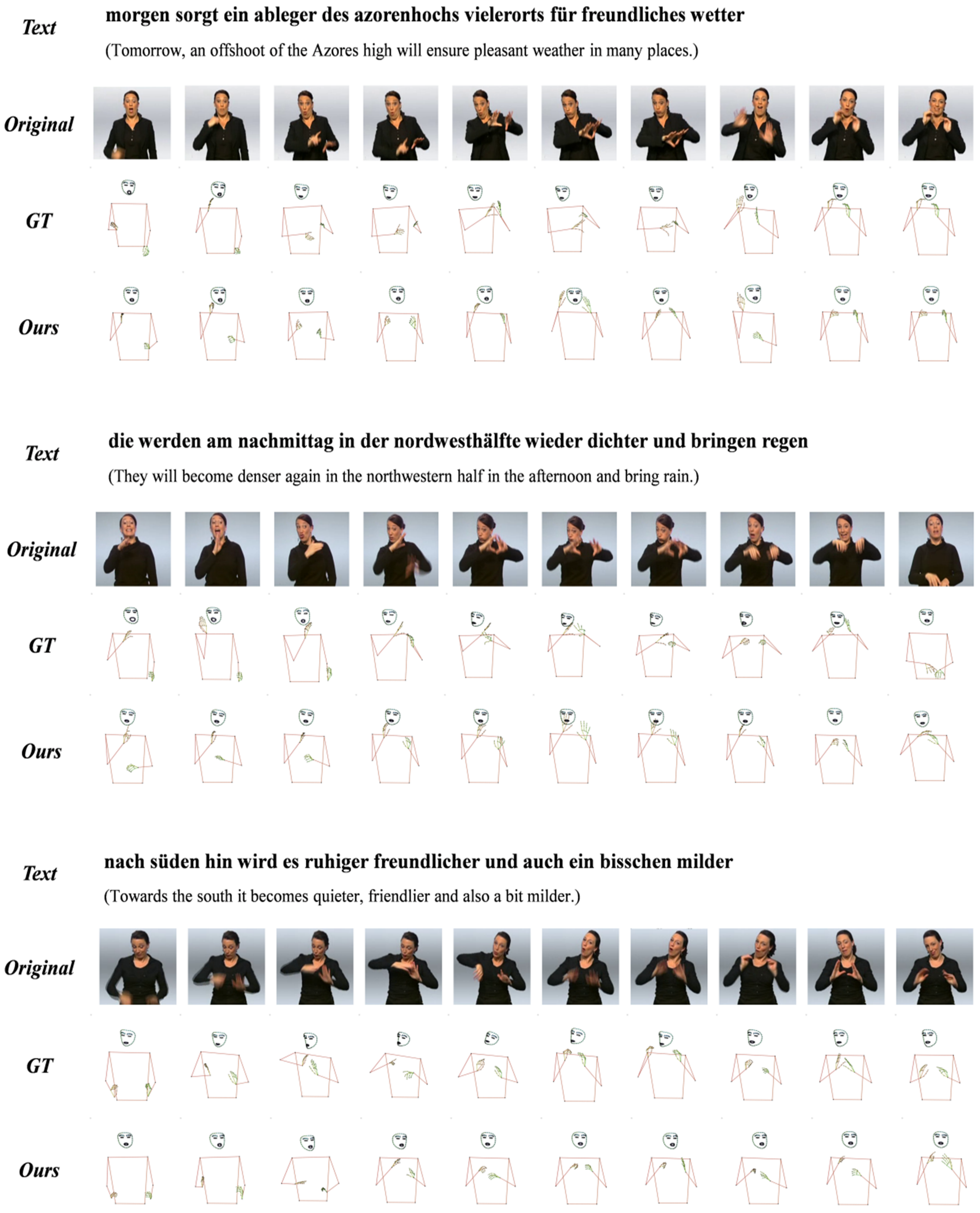}
  \caption{Visualization examples of generated pose sequences on PHOENIX14T.}
  \label{fig:Visual 2}
\end{figure*}

\subsection{Loss function} \label{Loss Function}
In our framework, we employ two distinct loss functions to boost the model's performance. Specifically, the $\mathcal{L}_{joint}$ is utilized to enforce the accuracy of the joints, while the $\mathcal{L}_{bone}$ serves to constrain the orientation of the bones.

Joint Constraint: To guarantee an exact alignment with the ground truth, we use the joint loss function to enforce the accuracy of joint positions in the pose. The joint constraint $\mathcal{L}_{joint}$ is defined as follows:
\begin{eqnarray}
    \mathcal{L}_{joint} = \frac{1}{J}\sum_{j=1}^{J}|p_{j}-p_{j}'|,
    \label{eq: Joint Constraint}
\end{eqnarray}
where $J$ represents the total number of joints. This formula measures the error in joint positions by computing the absolute difference between the predicted joint position $p_{j}'$ and the true joint position $p_{j}$, and then taking the average across all joints.

Bone Constraint: To enhance the accuracy of bone orientation in the generated poses, we introduce $\mathcal{L}_{bone}$. The formula is as follows:

\begin{eqnarray}
    \mathcal L_{bone} = \frac{1}{B}\sum_{b=1}^{B}(q_b-q_b')^2,
    \label{eq: Bone Constraint}
\end{eqnarray}
where $q_b$ and $q_b'$ represent the bone orientations obtained from the ground truth and the predicted results $p_{0}'$, and $B$ represents the number of bones. As shown in Figure~\ref{fig: Pose}, we calculate the direction of bone movement by dividing the parent and child nodes. The specific formula is as follows:
\begin{eqnarray}
    q=({\mathop{{x}_b}\limits ^{\rightarrow}}, {\mathop{{y}_b}\limits ^{\rightarrow}}, {\mathop{{z}_b}\limits ^{\rightarrow}}).
\end{eqnarray}
By computing the sum of squared differences between the actual and predicted bone directions and then averaging them, we can effectively quantify the error in bone orientation, thus enhancing the accuracy of bone orientation.

The final optimization goal is:
\begin{eqnarray}
    \mathcal{L} = \mathcal L_{joint} + \lambda \mathcal L_{bone},
\end{eqnarray}
where $\lambda$ is $0.1$. Please note that the $\mathcal{L}_{bone}$ is only applicable to the body and not to the facial area.

\begin{figure}[ht]
\centering
\includegraphics[width=0.98\columnwidth]{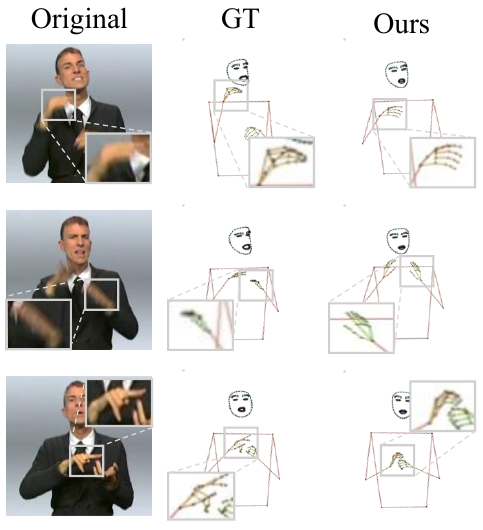}
\caption{Visualization examples of generated pose on PHOENIX14T.}
\label{fig:Visual 3}
\end{figure}
\subsection{Dataset}
During our experiments, the PHENIX14T dataset~\cite{Camgoz_Hadfield_Koller_Ney_Bowden_2018} acts as the main data source. This dataset is a comprehensive compilation, comprising 8,257 complete facial and gesture instances. These instances cover 2,887 unique German words and 1,066 various signs. This rich dataset offers a broad spectrum of data for our research, allowing us to carry out thorough investigations into the relationships among facial expressions, gestures, and the corresponding linguistic elements within the context of German sign language.

\subsection{Implementation Details}
Regarding the model architecture, the text encoder within our framework has the same structure as the Progressive Transformer~\cite{saunders2020progressive}. This choice is made to take advantage of the demonstrated effectiveness of the Progressive Transformer in processing sequential data, which is essential for encoding the text input of our model.

Concerning the model parameters, in our proposed method, both the text encoder and the denoiser $D$ are set up with 4 layers and 8 heads, sharing an embedding dimension of 1024. These parameter settings were meticulously chosen via preliminary experiments and analyses. The 4-layer architecture allows for an adequate depth of feature extraction, whereas the 8-head mechanism empowers the model to concurrently capture different aspects of the input data. An embedding dimension of 1024 offers a high-dimensional representation of the data, facilitating better encoding and subsequent processing.

Our diffusion model is equipped with a cosine scheduler. The scheduler sets $T = 1000$. The noise injection operations are crucial for the diffusion model to learn the data distribution in a noisy setting. Moreover, in the inference stage $i=5$. This number of inferences has been determined to strike a balance between computational cost and the accuracy of the generated outcomes.

During the training stage, we utilize the Adam optimizer~\cite{Kingma_Ba_2014}. The Adam optimizer is selected due to its efficiency and adaptability in dealing with various types of neural network models. It adjusts the learning rate of each parameter individually, which can result in quicker convergence. In our experiment, the learning rate is configured to $1 \times 10^{-3}$. This learning rate has proven to be effective in our experiments, enabling the model to converge smoothly without overshooting or being trapped in local minima.

All of our experiments are carried out using the PyTorch framework. The PyTorch framework offers a flexible and efficient platform for constructing and training neural network models. We employ an NVIDIA GeForce RTX 4090 GPU to conduct the training process.

\subsection{Experimental Results}
\subsubsection{Comparison with other teams}

As shown in Table~\ref{tab: result}, we provide a comprehensive report of the results achieved on the PHOENIX14T test set. This test set was part of a highly competitive challenge, in which numerous outstanding teams participated. Our team's performance, though it ended up in second place, has many remarkable aspects. Specifically, when evaluated using the DTW metric, our method showed its superiority. It not only surpassed the first-placed "USTC-MoE" team in this specific metric but also significantly outperformed all the other competing teams. This demonstrates the effectiveness and competitiveness of our approach, even though we didn't achieve the overall first place. The DTW metric, renowned for its importance in accurately measuring the similarity of time-series data, attests to the robustness and precision of our method when dealing with the complex data in the PHOENIX14T.

\subsubsection{Visualize results}

In the examples vividly depicted in Figures \ref{fig:Visual 1} and \ref{fig:Visual 2}, our proposed method demonstrates remarkable performance. As can be clearly observed in Figure \ref{fig:Visual 3}, when compared with the ground truth, the pose generated by our method bears a striking resemblance to the original image. This similarity is not merely a matter of general appearance but extends to the finest details. 

Throughout the entire process, our method showcases its prowess by effectively and precisely capturing every minute variation in hand movement. Starting from the initial, static gesture, transitioning through the complex and dynamic sequence of motions, and culminating in the final, often fleeting posture, the method leaves no nuance unaccounted for. 

When conducting a side-by-side comparison with the ground truth, it becomes patently clear that not only are the positions of the key nodes of the hand, such as the fingertips and knuckles, accurately pinpointed, but also the orientation of the overall hand contour is impeccably aligned. 
\section{Conclusions}
\label{sec:conclusions}
This research paper presents a novel solution to tackle the SLRTP Sign Production Challenge within the framework of CVPR 2025. Our proposed methodology builds on the Progressive Transformer baseline, which provides a robust foundation for sequence modeling. To further augment the model's capability in processing sign language sequences, we incorporate a diffusion model into the framework. This incorporation significantly enhances the model's capacity to capture the elaborate temporal and spatial inter-dependencies inherent in sign language data.
A series of extensive experiments were carried out, and our method manifested outstanding performance. Notably, on the PHOENIX14T dataset, it yielded an impressive experimental outcome of 20.17 BLEU-1, underscoring the efficacy of our proposed approach in accurately generating sign language representations. 

In future endeavors, we are dedicated to exploring more deeply into the application of diffusion models in sign language production. This exploration will entail examining various facets, such as optimizing the diffusion process to meet the sign language-specific needs, exploring different architectures of diffusion models that might be better adapted to this domain, and conducting more comprehensive assessments on a broader spectrum of datasets and tasks associated with sign language production.

\begin{table}[tbp]
\renewcommand\arraystretch{1.1}
\caption{The summary of final test set results.}
\label{tab: result}
\centering
\resizebox{\linewidth}{!}{
\begin{tabular}{cccccc}
\bottomrule[1pt]
   \multicolumn{1}{c}{Team} &BLEU-1{$\uparrow$} &BLEU-4{$\uparrow$} &ROUGE{$\uparrow$} &WER{$\downarrow$} &DTW{$\downarrow$}\\
\bottomrule[0.5pt]
   \multicolumn{1}{c}{USTC-MoE}  &21.35 &4.93  &23.26 &109.38 &0.0574\\
   \multicolumn{1}{c}{\textbf{hfum-lmc}}  &\textbf{20.17} &\textbf{4.44} &\textbf{22.20} &\textbf{107.93} &\textbf{0.0492}\\
   \multicolumn{1}{c}{Hacettepe}  &15.88  &2.41  &15.40 &105.49 &0.0531 \\
\bottomrule[1pt]
\end{tabular}}
\end{table}

{
    \small
    \bibliographystyle{ieeenat_fullname}
    \bibliography{main}
}


\end{document}